%% file: ijcai24.tex
\documentclass{article}
\usepackage{ijcai24}

\usepackage{times}
\usepackage{soul}
\usepackage{url}
\usepackage[colorlinks]{hyperref}
\usepackage[utf8]{inputenc}
\usepackage[small]{caption}
\usepackage{graphicx}
\usepackage{amsmath}
\usepackage{amsthm}
\usepackage{booktabs}
\usepackage{algorithm}
\usepackage{algorithmic}
\usepackage[switch]{lineno}
\title{Revealing the Two Sides of Data Augmentation: \\ 
An Asymmetric Distillation-based Win-Win Solution for Open-Set Recognition}

\author{
Yunbing Jia$^{1,2}$,
Xiaoyu Kong$^3$,
Fan Tang$^{2}$,
Yixing Gao$^{1,4}$\footnote{Corresponding author},
Weiming Dong$^5$,
Yi Yang $^6$\\
\affiliations
$^1$School of Artificial Intelligence, Jilin University $^2$Institute of Computing Technology, CAS \\
$^3$Harbin Institute of Technology (Shenzhen) 
$^4$Engineering Research Center of Knowledge-Driven Human-Machine Intelligence, Ministry of Education 
$^5$MAIS, Institute of Automation, CAS 
$^6$DIDI \\
\emails
jiayb22@mails.jlu.edu.cn,
21S151102@stu.hit.edu.cn,
tfan.108@gmail.com,\\
gaoyixing@jlu.edu.cn,
weiming.dong@ia.ac.cn,
yangyiian@didiglobal.com
}

\usepackage{xcolor}
\input{IJCAI_2024/macros}
\usepackage{diagbox}
\usepackage{multirow}
\usepackage{booktabs}
\usepackage{bbding}
\usepackage{booktabs}
\usepackage{array}
\usepackage{amsthm,amsmath,amssymb}
\usepackage{mathrsfs}
\usepackage{dsfont}

\usepackage{appendix}
\begin{document}

\maketitle

\begin{abstract}
    In this paper, we reveal the two sides of data augmentation: enhancements in closed-set recognition correlate with a significant decrease in open-set recognition.
    Through empirical investigation, we find that multi-sample-based augmentations would contribute to reducing feature discrimination, thereby diminishing the open-set criteria.
    Although knowledge distillation could impair the feature via imitation, the mixed feature with ambiguous semantics hinders the distillation.
    To this end, we propose an asymmetric distillation framework by feeding the teacher model extra raw data to enlarge the benefit of the teacher.
    Moreover, a joint mutual information loss and a selective relabel strategy are utilized to alleviate the influence of hard mixed samples.
    Our method successfully mitigates the decline in open-set and outperforms SOTAs by $2\%\sim3\%$ AUROC on the Tiny-ImageNet dataset, and experiments on large-scale dataset ImageNet-21K demonstrate the generalization of our method.
\end{abstract}

\input{AnonymousSubmission/Sections/1_Introduction}
% \input{AnonymousSubmission/Sections/2_RelatedWork}
\input{AnonymousSubmission/Sections/3_Analysis}
\input{AnonymousSubmission/Sections/4_Method}
\input{AnonymousSubmission/Sections/5_Experiments}
\input{AnonymousSubmission/Sections/6_Conclusion}
\newpage
\section*{Acknowledgments}
This work is supported in part by the Beijing Natural Science Foundation
under No. L221013, the National Natural Science Foundation of China under Grant Nos. 62102162 and 62203184, and the  CCF-DiDi GAIA Collaborative Research Funds for Young Scholars.

% \bibliographystyle{IJCAI_2024/named}
% \bibliography{IJCAI_2024/ijcai24.bib}

\newpage
% \includepdfmerge{IJCAI_2476_Paper_Supplementary.pdf,1-3}
\appendix

\section*{Supplementary Material}
\input{AnonymousSubmission/Sections/2_RelatedWork}

\input{IJCAI_2024/Sections/Analysis}
\input{IJCAI_2024/Sections/Extension}
\input{IJCAI_2024/Sections/Experiments}

\end{document}

%% file: IJCAI_2024/macros.tex
\usepackage{ifthen}

\newcommand{\warning}[1]{{\it\color{red} #1}}
\newcommand{\toremove}[1]{{\it\color{red} (To remove) #1}}
\newcommand{\note}[1]{{\it\color{blue} #1}}
\newcommand{\nothing}[1]{}

\definecolor{FanColor}{rgb}{0.8,0,0.8}

{

\renewcommand{\warning}[1]{}
\renewcommand{\toremove}[1]{}
\renewcommand{\note}[1]{}
\renewcommand{\nothing}[1]{}
}{}

\newcommand{\etal}{\textit{et al.}}

%% file: AnonymousSubmission/Sections/1_Introduction.tex
%引言
\section{Introduction}
\label{sec:introduction}
    The utilization of data augmentation (DA) strategies in training neural networks have been proven effective in expanding the training dataset~\cite{yang2022image} and have become widespread in many applications~\cite{chen2021siamcpn,xu2022transformers,chen2023spa,hou2024augmentation,wang2024class}.
    \input{AnonymousSubmission/Figures/fig_1}
    As the simplest implementation, the base manipulation-based DA is the most common strategy and can be divided into two categories: single-sample-based augmentation (SSA) and multiple-sample-based augmentation (MSA).
    SSA creates new samples by conducting basic operations on a single sample, including rotation, flipping, blurring, or their combinations~\cite{devries2017cutout,cubuk2018autoaugment,hendrycks2020augmix}.
    Meanwhile, MSA further increases the diversity by involving more than one sample to generate the convex combination of them, i.e., cut-and-paste or addition~\cite{zhang2017mixup,yun2019cutmix} and hence remarkably boosts the closed-set recognition ability as shown in Figure~\ref{fig:fig_1}.

    Efficient and effective as MSA is, some research found it affects the performance of recognition tasks to some extent. Balestriero~\etal~\shortcite{balestriero2022effects} demonstrate that MSA caused a drop in some classes, and Choi~\etal~\shortcite{choi2023understanding} argue that MSA disperses features in similar classes.
    However, compared with closed-set recognition, open-set recognition (OSR) is actually the biggest victim of this problem because it has no access to open-set data and hence heavily relies on the discriminative feature. 
    As shown in Figure~\ref{fig:fig_1}, we reveal that the significant improvement of MSA on closed-set recognition sacrifices the performance of OSR, and as closed-set recognition improves, the corresponding decline in OSR becomes more pronounced, dubbed as the two sides of MSA.
    
    To mitigate the degradation of open-set performance caused by MSA, Roady~\etal~\shortcite{roady2020improved} tempered the outputs of the models in a label-smoothing way to increase the entropies of the model's outputs.
    Xu~\etal~\shortcite{xu2023contrastive} implemented the InfoNCE loss and used MixUp to enlarge the inter-class margins.
    However, these extra constraints alleviate the dilemma of MSA on OSR while ceiling the improvement on closed-set recognition.
    We argue that an ideal solution should be win-win for both closed- and open-set samples.
    
    Based on preliminary experiments on the interplay of DA and OSR, we have two key observations:
    1) \textit{MSA performs worse than SSA on OSR since it would disperse the features};
    2) \textit{Knowledge distillation benefits OSR but goes back to decline when MSA joins in}.
    Digging deeper into these observations, we found that MSA diminishes the criteria of OSR in two aspects.
    First, MSA degrades the magnitude of the activation of features and logits, which leads to great uncertainty in selecting unknown samples via the logits threshold.  
    Distillation mitigates this problem somewhat by forcing the student network to mimic the activation magnitude of the teacher network. Secondly, low-discriminative features of MSA samples remain uncertain; merely distillating them still suffers OSR criteria diminution.
    
    Motivated by the above observation and findings, we propose an asymmetric distillation framework, a win-win solution for both close-set and open-set performance.
    Concretely, in addition to the same MSA samples fed to the teacher and student in symmetric distillation, we introduce extra raw samples to the teacher and exert extra mutual information objectively to enlarge the teacher's benefit.
    The introduced objective enables the student to focus more on the class-specific features within the mixed samples.
    Moreover, since some hard mixed samples provide ambiguous semantic information, we filter them out by relaxedly checking the teacher's predictions and assigning them an unknown-like target to encourage the model to decrease its activation for the non-salient features of the known classes.
    Within this framework, the model can leverage the advantages of MSA on closed-set performance and better discriminate the novels under extra supervision.
    
    The main contributions in this paper are as follows:
    \begin{itemize}
    \item We revealed the two sides of DA leading to the degeneration of OSR and conducted experiments to analyze how the augmented samples undermine the model.

    \item We introduce an asymmetric distillation framework with a cross-mutual information maximization and a two-hot label smoothing to eliminate the effect and further improve the model's open-set performance.

    \item We perform extensive experiments and prove the effectiveness of our proposed method on various benchmarks.
    \end{itemize}

%% file: AnonymousSubmission/Figures/fig_1.tex
\begin{figure}[!t]
\centering
\includegraphics[width=\linewidth]{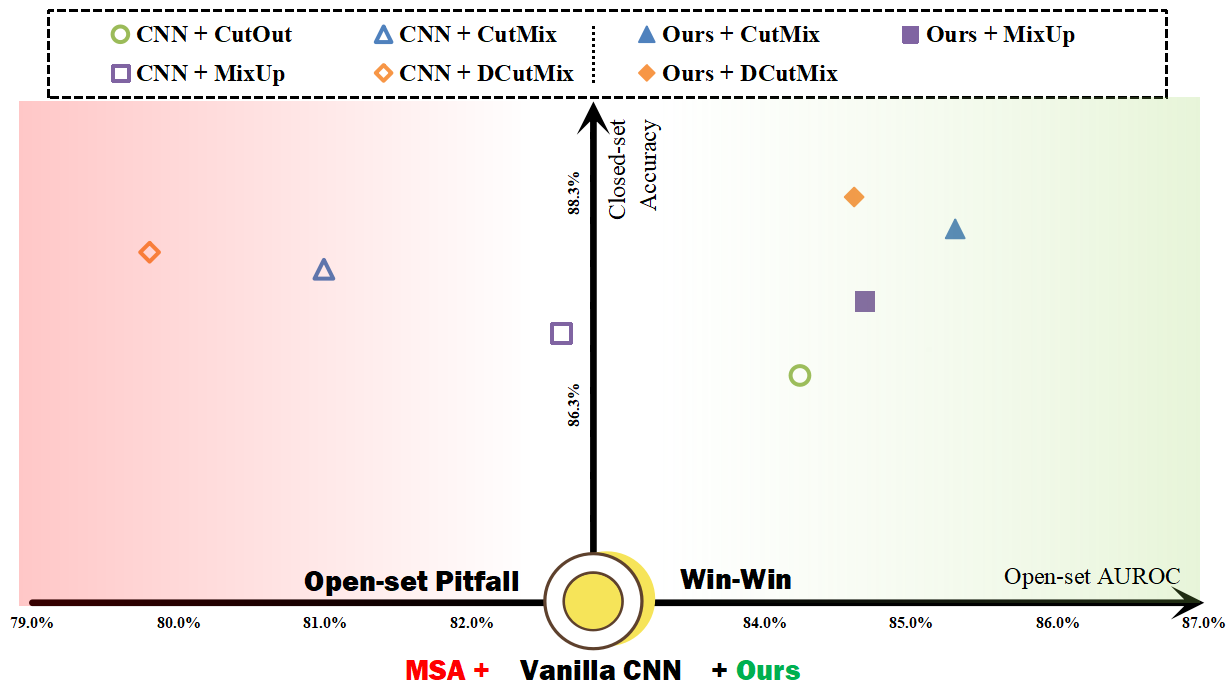}
%\vskip -0.5em
\caption{Illustration of the two sides of data augmentation. Despite the tremendous accuracy gain made by augmentations, multiple sample-based augmentation (MSA) tends to degrade the model's open-set performance.} 
\label{fig:fig_1}
\end{figure}

%% file: AnonymousSubmission/Sections/3_Analysis.tex
%分析
\section{Reveal the Two Sides of MSA}
\label{sec:Analysis}
    Despite MSA achieves significant improvement in closed-set recognition, we reveal two sides of MSA on closed-set and OSR in this section.
    We first present two distinct observations concerning DA and OSR in Section~\ref{sec:3.1}, followed by an in-depth analysis expounded in Section~\ref{sec:visualize}.
    Finally, we elucidate the mechanisms through which knowledge distillation can alleviate the degradation in OSR performance induced by MSA, and we highlight inherent issues within existing symmetric distillation frameworks in Section~\ref{sec:distillation}.

    \input{AnonymousSubmission/Tables/mix_out_compare}
    \input{AnonymousSubmission/Figures/fig_union}
    \subsection{Key Findings from DA and OSR Interplay}
    \label{sec:3.1}
    Without loss of generality, we experiment on both SSA and MSA methods on accuracy and Area Under the Receiver Operating Characteristic curve (AUROC) on Tiny-ImageNet dataset.
    % We choose three SSA methods: CutOut~\cite{devries2017cutout} randomly masks a region of an image.
    % Randomized quantization~\cite{wu2023randomized} is a data-agnostic augmentation based on quantization along the channel dimension.
    % Augmix~\cite{hendrycks2020augmix} leverages simple augmentation operations in concert with a consistency loss.
    % For MSA, we take the commonly used MixUp~\cite{zhang2017mixup} and its variant CutMix~\cite{yun2019cutmix} which cuts a small patch from a source image and pastes it to another source image for example.
    As shown in Table~\ref{tab:mix_out_compare}, taking the vanilla model as the baseline, we can make the following two observations: 

    \textbf{Observation 1) on SSA vs MSA.} % 就着表说
    Both SSA and MSA exhibit efficacy in enhancing the closed-set accuracy of the model, attributed to their capabilities in expanding the dataset. 
    Notably, MSA demonstrates superior performance in this regard, as it effectively enlarges the diversity within the training data.
    Nevertheless, when evaluating AUROC, SSA modestly enhances performance, while the incorporation of MSA significantly undermines OSR capabilities.
    
    \textbf{Observation 2) Distillation Benefits OSR.} % distillation / logits representation
    To verify the influence of distillation on MSA, we compare the MSA-sample based distillation with vanilla distillation framework.
    Following Wang~\etal~\shortcite{wang2022what}, we use a non-MSA-trained network as the teacher model. 
    As shown in Table~\ref{tab:mix_out_compare}, with distillation only, both accuracy and AUROC gain an improvement.
    Furthermore, the integration of MSA significantly enhances accuracy by a substantial margin. 
    This implies that the dataset expanded through MSA contributes significantly to the augmentation of the model's representational capacity. 
    Notwithstanding the attainment of more expressive and generalized features, MSA persists in yielding a decrement rather than an amelioration in performance on OSR.

    \subsection{MSA Diminishes the Criteria of OSR}
    \label{sec:visualize}
    Choi~\etal~\shortcite{choi2023understanding} argue that the MixUp-trained model disperses features in closed-set classes.
    To quantify such a phenomenon, we visualize the discrepancy among all class pairings of the vanilla model and the CutMix-trained model in Figure~\ref{fig:union} (a).
    At first glance, the heatmap of the CutMix-trained model is darker than the vanilla one, which indicates the degradation of the model's activation magnitude and the lower margins among all the classes.
    Specifically, the drastic decrease of the gap among the similar classes `k\_2 - k\_3' and `k\_2 - k\_5' in Figure~\ref{fig:union} (a) indicates that the model tends to learn an obscure boundary among these classes.
    In contrast, the degradation of the distinct classes such as `k\_4 - k\_5' and `k\_4 - k\_3' is slighter.
    The broken boundaries among the similar classes are vulnerable to the unknowns which have similar features with these classes.
    For OSR, the darker colors in the intersect regions of the known classes and the unknown classes suggest the model's degeneration of discriminating them from each other.
    To dig deeper into the observation, we draw a theoretical analysis in the following.

    Denoting $D = \left\{D_{k}, D_{unk} \right\} $ as all the inputs the model may encounter during deployment, $ D_{train} = \left\{ {\text{x}_i , \text{y}_i} \right\}_{i=1}^{n} \subseteq D_{k} $ represents the training dataset and $\text{x}_i$ and $ \text{y}_i$ are the image and the corresponding label.
    Given input image $\text{x}_i$, model's $C$-classes prediction  $\hat{\text{y}}_i$ can be obtained via $\hat{\text{y}}_i = \text{softmax}(\text{\textbf{W}}\Phi_{\theta}(\text{x}_i))$, where $\Phi_{\theta}(\cdot)$ is the feature extractor and $\Phi_{\theta}(\text{x}_i) \in \mathbb{R}^D$. 
    $\textbf{W} = [\textbf{w}_1, \textbf{w}_2, \dots, \textbf{w}_C] \in \mathbb{R}^{C \times D}$ is the linear classification matrix and $\text{\textbf{W}}\Phi_{\theta}(\text{x}_i)=[\hat{y}_{i, 1},\hat{y}_{i, 2},\dots,\hat{y}_{i, C}]$ is the logits.
    The training is based on the cross-entropy loss $\mathcal{L}_{CE}$:
    \begin{equation}
    \begin{split}
        \mathcal{L}_{CE}&(\theta, \textbf{W})  = -\hat{y}_{i,c} + \text{log}(\sum_{k=1}^C \text{exp}(\hat{y}_{i, k})) \\
        & = -\textbf{w}_c\,\Phi_{\theta}(\text{x}_i) + \text{log}(\sum_{k=1}^C \text{exp}(\textbf{w}_k \, \Phi_{\theta}(\text{x}_i))).
    \label{eq:ce_loss}
    \end{split}
    \end{equation}
    
    Vaze~\etal~\shortcite{vaze2022openset} investigated how $\mathcal{L}_{CE}$ influences OSR.
    The model initially embeds all the classes with a similar magnitude and gradually activates more for the known classes by increasing $||\textbf{W}\,\Phi_{\theta}(\text{x})||$ to better distinguish the unknowns.
    The final maximum logit score is used to provide the open-set score in their conclusion.
    Additionally, the model's wrong predictions during training tend to reduce $\textbf{w}_k \cdot \Phi_{\theta} (\text{x}_i) \,\forall k \neq c$.

    We use CutMix as an example to study the impact of MSA:
    \begin{equation}
    \begin{split}
        &\text{x}_m = \textbf{M} \odot \text{x}_i + ( \textbf{1} - \textbf{M}) \odot \text{x}_j,\\
        &\text{y}_m = \lambda \cdot \text{y}_i + (1 - \lambda) \cdot \text{y}_j,
    \label{eq:mix_generate}
    \end{split}
    \end{equation}
    where \textbf{M} is a mask and $\lambda$ is sampled from Beta distribution $\beta(\alpha, \alpha)$.
    % Following Yun~\etal~\shortcite{yun2019cutmix}, we set $\alpha$ to 1.0 in the experiments.
    With $\text{x}_m$, Eq.~\ref{eq:ce_loss} can be rewritten as:
    \begin{equation*}
        \mathcal{L}_m(\theta, \textbf{W}) = (-\lambda \cdot \textbf{w}_{c_1} \, \Phi_{\theta}(\text{x}_m)) + (-(1 - \lambda) \cdot \textbf{w}_{c_2} \, \Phi_{\theta}(\text{x}_m)) 
    \end{equation*}
    \begin{equation}
         + \text{log}(\sum_{k=1}^C \text{exp}(\textbf{w}_k \, \Phi_{\theta}(\text{x}_m))),
    \label{eq:mix_loss}
    \end{equation}
    where $c_1$ and $c_2$ are the ground-truth classes of $\text{x}_i$ and $\text{x}_j$.

    We display the comparison of $||\Phi_{\theta}(\text{x})||$ and $||\textbf{W}\,\Phi_{\theta}(\text{x})||$ in Figure~\ref{fig:union} (b) to explore how Eq.~\ref{eq:mix_loss} influences the model's behavior.
    It is straightforward that the MSA-trained model suffers from a degradation of feature norm which have the direct bearing on the model's criteria of OSR.
    Consequently, in Figure~\ref{fig:union} (b), $||\textbf{W}\,\Phi_{\theta}(\text{x})||$ of the MSA-trained model decreases drastically, thus harming the open-set score.
    The discrepancy between the known classes and the unknown classes is also reduced as can be seen in Figure~\ref{fig:union} (b).
    Through the above analysis, we conclude that MSA diminishes the criteria of OSR.
    
    \subsection{Retrieve the Discrepancy by Distillation}
    \label{sec:distillation}
    The distillation experiments in Table~\ref{tab:mix_out_compare} indicate that KD benefits OSR.
    However, distillation with CutMix brings a greater improvement on the model's accuracy while impairing the gain of OSR performance.
    We investigate how the teacher works by analyzing the distillation loss $\mathcal{L}_{Distill} = \mathcal{D}_{KL} (\hat{\text{y}}^s|| \hat{\text{y}}^t)$, where the superscripts $s$ and $t$ denote the student and the teacher model.
    It encourages $\hat{\text{y}}^s$ to minimize its divergence with $\hat{\text{y}}^t$, which implicitly leads to an alignment of the magnitude of the activation.
    This can be concluded in Figure~\ref{fig:union} (b) by comparing the $||\Phi_{\theta}(\text{x})||$ of the KD-trained model and the teacher.
    In addition, the comparison of $||\textbf{W}\,\Phi_{\theta}(\text{x})||$ between the MSA-trained model and the MSA-distilled model indicates that distillation with MSA helps the model retrieve the decreased discrepancy between the known classes and the unknown classes to a certain degree.
    However, the MSA is still harmful to the distilled model, which suggests that the vanilla symmetric distillation framework can not mitigate this issue.

    To investigate why CutMix influences the benefit of distillation, we calculate the teacher's over-confident predictions (the maximum probability is greater than 95\%) and wrong predictions (the predicted class is out of $c_1$ and $c_2$) over 10000 mixed samples on Tiny-ImageNet dataset as shown in Figure~\ref{fig:union} (c).
    The statistical results suggest that the teacher makes amount of unreasonable predictions on MSA samples.
    For example, the mixtures of the similar classes will easily be over-confidently predicted because of the redundant activation of their similar features.
    To solve this problem, We regularize the teacher's redundant activation by an asymmetric distillation framework with an extra mutual information supervision and a re-label mechanism in Section~\ref{sec:method_m}.
    And the wrong prediction indicates that the mixed sample does not include the discriminative features so that the model should be encouraged to decrease its activation.
    To achieve this, we re-label the wrong predicted samples with smoothed two-hot labels to make the model put less attention on the class-agnostic features within the mixed samples.

%% file: AnonymousSubmission/Tables/mix_out_compare.tex
% \usepackage{multirow}
\begin{table*}[h]
\renewcommand\arraystretch{1.3}
\centering
\caption{
The impact of different augmentations on different models. 
We report the close-set accuracy (Acc.) and AUROC.
The green numbers in the upper right show the improvement compared to the vanilla CNN model and the numbers in red indicate the degradation.
}
%\vskip -0.5em
\label{tab:mix_out_compare}
\scalebox{0.54}{
\begin{tabular}{lcc|cc|cc|cc|cc|cc|cc|cc}
\toprule[2pt]
\multirow{3}{*}{Model} & \multicolumn{2}{c|}{\multirow{2}{*}{Vanilla CNN}} & \multicolumn{6}{c|}{SSA} & \multicolumn{4}{c|}{MSA} & \multicolumn{4}{c}{Distillation}  \\ \cline{4-17}

& & & \multicolumn{2}{c|}{+ AugMix~\shortcite{hendrycks2020augmix}} & \multicolumn{2}{c|}{+ Rand. Quantization~\shortcite{wu2023randomized}} & \multicolumn{2}{c|}{+ CutOut~\shortcite{devries2017cutout}} & \multicolumn{2}{c|}{+ CutMix~\shortcite{yun2019cutmix}} & \multicolumn{2}{c|}{+ MixUp~\shortcite{zhang2017mixup}} & \multicolumn{2}{c|}{Vanilla} & \multicolumn{2}{c}{+ CutMix} \\ \cline{2-17} 
                       & Acc. & AUROC    & Acc. & AUROC    & Acc. & AUROC    & Acc. & AUROC    & Acc. & AUROC    & Acc. & AUROC    & Acc. & AUROC    & Acc. & AUROC \\ \hline
ResNet-101             & $86.72$ & $84.03$    & $86.98^{\textcolor{green}{+0.26}}$ & $84.12^{\textcolor{green}{+0.09}}$    & $86.90^{\textcolor{green}{+0.18}}$ & $84.14^{\textcolor{green}{+0.11}}$    & $86.84^{\textcolor{green}{+0.12}}$ & $84.11^{\textcolor{green}{+0.08}}$    & $87.06^{\textcolor{green}{+0.34}}$ & $82.62^{\textcolor{red}{-1.41}}$    & $88.34^{\textcolor{green}{+1.62}}$ & $83.74^{\textcolor{red}{-0.29}}$    & $87.32^{\textcolor{green}{+0.60}}$ & $84.65^{\textcolor{green}{+0.62}}$    & $88.90^{\textcolor{green}{+2.18}}$ & $84.50^{\textcolor{green}{+0.47}}$ \\
ResNet-50              & $86.16$ & $83.84$    & $86.28^{\textcolor{green}{+0.12}}$ & $83.92^{\textcolor{green}{+0.08}}$    & $86.42^{\textcolor{green}{+0.26}}$ & $84.19^{\textcolor{green}{+0.32}}$    & $86.36^{\textcolor{green}{+0.20}}$ & $84.26^{\textcolor{green}{+0.42}}$    & $87.44^{\textcolor{green}{+1.28}}$ & $83.41^{\textcolor{red}{-0.43}}$    & $86.62^{\textcolor{green}{+0.46}}$ & $83.13^{\textcolor{red}{-0.71}}$    & $86.38^{\textcolor{green}{+0.22}}$ & $84.74^{\textcolor{green}{+0.90}}$    & $88.04^{\textcolor{green}{+1.88}}$ & $84.10^{\textcolor{green}{+0.26}}$ \\
ResNet-18              & $84.28$ & $82.84$    & $86.14^{\textcolor{green}{+1.86}}$ & $84.10^{\textcolor{green}{+1.26}}$    & $84.64^{\textcolor{green}{+0.36}}$ & $82.93^{\textcolor{green}{+0.09}}$    & $86.42^{\textcolor{green}{+2.14}}$ & $84.24^{\textcolor{green}{+1.44}}$    & $87.42^{\textcolor{green}{+3.14}}$ & $80.99^{\textcolor{red}{-1.29}}$    & $86.82^{\textcolor{green}{+2.54}}$ & $82.61^{\textcolor{red}{-0.23}}$    & $86.64^{\textcolor{green}{+2.36}}$ & $84.24^{\textcolor{green}{+2.44}}$    & $88.82^{\textcolor{green}{+4.54}}$ & $82.47^{\textcolor{red}{-0.37}}$ \\
VGG-19                 & $82.10$ & $80.99$    & $82.70^{\textcolor{green}{+0.60}}$ & $81.69^{\textcolor{green}{+0.70}}$   & $82.86^{\textcolor{green}{+0.76}}$ & $81.32^{\textcolor{green}{+0.33}}$    & $82.40^{\textcolor{green}{+0.30}}$ & $81.17^{\textcolor{green}{+0.18}}$    & $83.44^{\textcolor{green}{+1.34}}$ & $75.24^{\textcolor{red}{-5.75}}$    & $83.34^{\textcolor{green}{+1.24}}$ & $76.81^{\textcolor{red}{-4.18}}$    & $83.12^{\textcolor{green}{+1.02}}$ & $81.18^{\textcolor{green}{+0.19}}$    & $84.18^{\textcolor{green}{+2.08}}$ & $76.72^{\textcolor{red}{-4.27}}$ \\
VGG-16                 & $80.90$ & $80.83$    & $82.84^{\textcolor{green}{+1.94}}$ & $81.51^{\textcolor{green}{+0.68}}$    & $82.96^{\textcolor{green}{+2.06}}$ & $81.86^{\textcolor{green}{+1.03}}$    & $82.36^{\textcolor{green}{+1.46}}$ & $81.17^{\textcolor{green}{+0.34}}$    & $84.28^{\textcolor{green}{+3.38}}$ & $74.98^{\textcolor{red}{-5.85}}$    & $83.10^{\textcolor{green}{+2.20}}$ & $75.78^{\textcolor{red}{-4.95}}$    & $83.74^{\textcolor{green}{+2.84}}$ & $81.62^{\textcolor{green}{+0.72}}$    & $85.04^{\textcolor{green}{+4.14}}$ & $78.17^{\textcolor{red}{-2.66}}$ \\
VGG-13                 & $80.72$ & $80.49$    & $83.86^{\textcolor{green}{+3.14}}$ & $81.70^{\textcolor{green}{+1.21}}$    & $82.18^{\textcolor{green}{+1.46}}$ & $81.67^{\textcolor{green}{+1.18}}$    & $82.98^{\textcolor{green}{+2.26}}$ & ${81.45}^{\textcolor{green}{+2.26}}$    & $83.96^{\textcolor{green}{+3.24}}$ & $74.90^{\textcolor{red}{-5.59}}$    & $83.08^{\textcolor{green}{+2.36}}$ & $73.00^{\textcolor{red}{-7.49}}$    & $83.62^{\textcolor{green}{+2.90}}$ & $81.80^{\textcolor{green}{+1.31}}$    & $85.32^{\textcolor{green}{+4.60}}$ & $78.87^{\textcolor{red}{-1.62}}$ \\
MobileNetV2            & $83.20$ & $81.31$    & $84.46^{\textcolor{green}{+1.26}}$ & $81.56^{\textcolor{green}{+0.25}}$    & $83.50^{\textcolor{green}{+0.30}}$ & $81.52^{\textcolor{green}{+0.21}}$    & $84.24^{\textcolor{green}{+1.04}}$ & $81.97^{\textcolor{green}{+0.66}}$    & $86.26^{\textcolor{green}{+3.06}}$  & $78.82^{\textcolor{red}{-2.49}}$    & $85.42^{\textcolor{green}{+2.22}}$ & $78.68^{\textcolor{red}{-2.63}}$    & $84.42^{\textcolor{green}{+1.22}}$  & $82.36^{\textcolor{green}{+1.05}}$    & $84.46^{\textcolor{green}{+1.26}}$ & $82.00^{\textcolor{green}{+0.69}}$ \\ \bottomrule[2pt]
\end{tabular}}

\end{table*}

%% file: AnonymousSubmission/Figures/fig_union.tex
\begin{figure*}[!t]
\centering
\includegraphics[width=\linewidth]{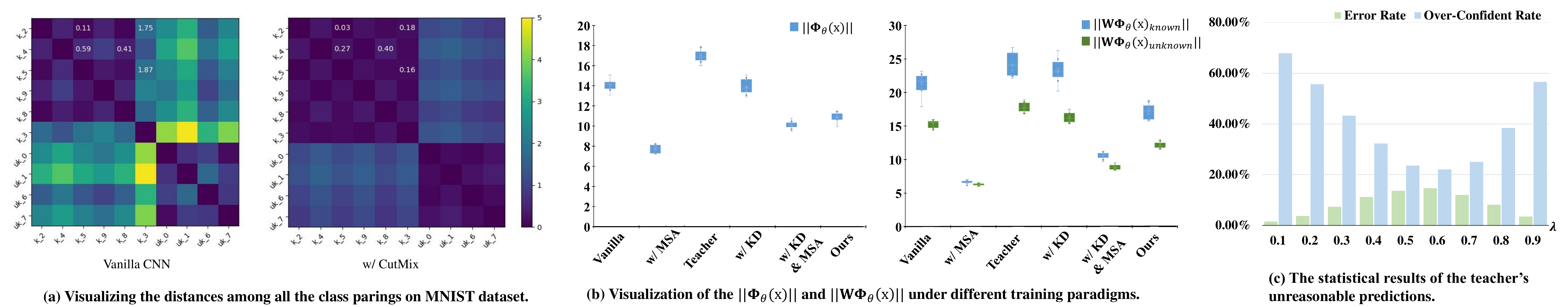}
\caption{(a) Heatmap visualization of the distances among all the class parings on MNIST dataset. `$k$' denotes the known classes and `$uk$' denotes the unknown classes. The number after the underline is the ground-truth label. (b) The comparison of $||\Phi_\theta(\text{x})||$ and $||\textbf{W}\,\Phi_\theta(\text{x})||$ under different training paradigms. (c) The teacher's top-2 error rate and over-confident predictions (higher than 95\%) over 10000 mixed samples under different mixing coefficients.}
\label{fig:union}
\end{figure*}

%% file: AnonymousSubmission/Sections/4_Method.tex
%方法
\section{Method}
\label{sec:method_m}
    \subsection{Overview}
        The overall pipeline of the proposed asymmetric distillation framework is outlined in Figure~\ref{fig:pipeline_new}.
        Based on the vanilla symmetric distillation in which the student and the teacher are fed with the same inputs, we introduce extra initial samples $\text{x}_i$ and $\text{x}_j$ to the teacher while training with the augmented input $\text{x}_m$ to perform an asymmetric data flow between the student and the teacher.
        We utilize the teacher's output of $\text{x}_i$ and $\text{x}_j$ to exert the student's output of $\text{x}_m$ a cross mutual information objective which forces the student to concentrate more on the class-specific features within $\text{x}_m$.
        In addition, for the confusing mixtures that are wrongly predicted by the teacher, we pick them out and re-label them with a smoothed two-hot label to decrease the student's activation of them.
        Achieving this can make the student less active in the class-agnostic features.

    \subsection{Asymmetric Distillation Framework}
        The asymmetric distillation framework is specially designed for the training of MSA.
        In our experiments, the model is randomly trained using either the original or mixed samples, with a probability of 0.5.
        We control different data flows with different objects for the student and the teacher during distillation to leverage the teacher's prior knowledge.

        \textbf{Training with the Initial Samples.}
        We inherit the advantages of KD for the initial samples by training with the symmetric distillation framework similar to Wang~\etal~\shortcite{wang2022what}.
        The loss is computed by the CE Loss and the Distillation Loss:
        \begin{equation}
            \mathcal{L}_{raw} = \mathcal{L}_{CE}(\hat{\text{y}}^s, \text{y}) + \mathcal{L}_{Distill}(\hat{\text{y}}^s, \hat{\text{y}}^t).
        \end{equation}

        \textbf{Asymmetric Inputs of Training MSA.}
        The unreasonable output of the teacher emphasizes the non-salient features within the mixture.
        To encounter this and enable the student to concentrate more on the class-specific features, we propose extra supervision to amplify the teacher's optimization of $||\Phi_\theta(\text{x})||$.
        Concretely, we build an asymmetric distillation framework upon the vanilla symmetric distillation framework by introducing the initial samples to the teacher.
        The additional initial samples are used to offer the mixed samples an extra mutual information maximization objective to amplify the teacher's impact.
        
        \input{AnonymousSubmission/Figures/fig_pipleine_new}
        \textbf{Cross Mutual Information.} Mutual Information (MI) is a fundamental measurement to quantify the relationship between random variables~\cite{hou2021disentangled,feng2023mutual} denoting by $\mathcal{I}\,(v_1, v_2)$ where $v_1$ and $v_2$ are two random variables.
        Initially, the primary objective for MSA training can be understood as maximizing $\mathcal{I}\,(\Phi_\theta^s(\text{x}_m), \,\text{y}_m)$.
        As the mixed label $\text{y}_m$ does not reflect the amount of the label information included in $\text{x}_m$ well, the powerful teacher is introduced to produce the embedding $\Phi_\theta^t(\text{x}_m)$ which is considered as a proper and model-friendly target of optimization.
        The objective of distillation can be abstracted into the maximization of $\mathcal{I}\,(\Phi_\theta^t(\text{x}_m), \,\Phi_\theta^s(\text{x}_m))$ which encourages the student to align its representation with the teacher.
        
        However, since $\Phi_\theta^t(\text{x}_m)$ may include class-agnostic information, especially the mixture of similar classes, we argue that merely imitating the teacher's output is not an optimal solution for OSR.
        An ideal objective encourages the student to maximize the discriminative features while discarding the common ones within the mixed samples.
        We achieve this by excluding their shared label information to amplify the teacher's impact on the class-specific features.
        The term $\mathcal{I}\,(\Phi_\theta^s(\text{x}_m), \,\Phi_\theta^t(\text{x}_m)|\text{y}_j)$ rigorously quantifies the amount of information of the $c_1$-th class shared between $\Phi_\theta^t(\text{x}_m)$ and $\Phi_\theta^s(\text{x}_m)$ where `$|$' is an excluding operation.
        The maximization of this term forces the student to attend more on the characteristic features of the $c_1$-th class in $\text{x}_m$.
        We maximize this term for both of the two classes in $\text{x}_m$ by a mutual information loss:
        \begin{equation}
        \begin{split}
            \mathcal{L}_{MI} = -(\,\mathcal{I}(\Phi_\theta^s(\text{x}_m), &\Phi_\theta^t(\text{x}_m)|\text{y}_j) \\
             &+ \mathcal{I}(\Phi_\theta^s(\text{x}_m), \Phi_\theta^t(\text{x}_m)|\text{y}_i)).% $}
        \label{eq:MI}
        \end{split}
        \end{equation}

        Revisiting the terms in Eq.~\ref{eq:MI}, we find that excluding the class-agnostic information of the $c_1$-th class in $\text{x}_m$, i.e. $(\Phi_\theta^t(\text{x}_m)|\text{y}_j)$, can be easily achieved by replacing it with $\Phi_\theta^t(\text{x}_i)$ in consideration of $\text{x}_i$ shares the same pure information of the $c_1$-th class with $\text{x}_m$.
        So we maximize the mutual information of $\text{x}_m$, $\text{x}_i$ and $\text{x}_j$ in a cross manner and simplify Eq.~\ref{eq:MI} to a Cross Mutual Information loss:
        \begin{equation}
        \begin{split}
            \mathcal{L}_{CMI} =  - (\lambda \mathcal{I}\,(\Phi_\theta^s(\text{x}_m), &\Phi_\theta^t(\text{x}_i)) \\
            & + (1 - \lambda)\mathcal{I}(\Phi_\theta^s(\text{x}_m), \Phi_\theta^t(\text{x}_j)),
        \label{eq:CMI}
        \end{split}
        \end{equation}
        where $\lambda$ is determined by Eq.~\ref{eq:mix_generate} to weight the contribution of $x_i$ and $x_j$.
    \subsection{Sample Verify and Two-Hot Label Smoothing}
    \label{sec:smoothed}
        Eq.~\ref{eq:CMI} enables the model to discard the class-agnostic features in $\text{x}_m$.
        However, in Figure~\ref{fig:union} (c), we point out that the teacher makes mistakes for some corner cases, which may be sub-optimal to the model's optimization.
        We revise the teacher's wrong predictions with a smoothed two-hot label to help the model learn more uncertainties within the confusing samples.

        \textbf{Relaxed Sample Verify.}
        For a mixed sample $\text{x}_m$, we argue that it contains the non-salient parts of both the $c_1$-th and the $c_2$-th class when the teacher predicts it to the third class.
        We utilize a relaxed verification that checks the top-2 accuracy of $\hat{\text{y}}^t_m$ to filter the teacher's wrong predictions out and assign them a revised target to optimize.

        \textbf{Two-Hot Label Smoothing.}
        We aim to optimize the wrongly predicted mixed samples to decrease their activation so that the model can discard the non-discriminative features within the mixtures.
        The cross-entropy loss we discussed above can naturally degrade the activation for the wrong predictions by Eq.~\ref{eq:ce_loss}.
        In addition, we want the model to learn more uncertainties among the confusing mixtures, so we manually set a revised target for these samples.
        Concretely, we mix a uniform label $\bar{\text{y}} \in \mathbb{R}^C$ whose elements are $1/C$ and $\text{y}_m$ by a ratio of 0.5 to generate the target of the wrongly predicted $\text{x}_m$ which we name it $\text{y}_u$.
        The loss is computed by:
        \begin{equation}
        \begin{split}
            \mathcal{L}_{Relabel} = &\mathds{1}(\text{argmax}(\hat{\text{y}}^t_m) \neq c_1\,and\,\text{argmax}(\hat{\text{y}}^t_m) \neq c_2)\\
            &\mathcal{L}_{CE}(\hat{\text{y}}^s_m, \text{y}_u),
        \end{split}
        \end{equation}
        where $\mathds{1}(\cdot)$ is an indicator function whose value is 1 when the following expression in the brackets is true and 0 vise versa.
        This object encourages the model to embed the uncertain samples $\text{x}_m$ to the origin of the feature space.

        The overall loss can be denoted as:
        \begin{equation}
            \mathcal{L}_{CutMix} = \mathcal{L}_{Distill} + \mu\,\mathcal{L}_{CMI} + \eta\,\mathcal{L}_{Relabel},
        \end{equation}
        where $\mu$ and $\eta$ are hyper-parameters we set to 1.0.

%% file: AnonymousSubmission/Figures/fig_pipleine_new.tex
\begin{figure}[!t]
\centering
\scalebox{1.0}{
\includegraphics[width=1.0\linewidth]{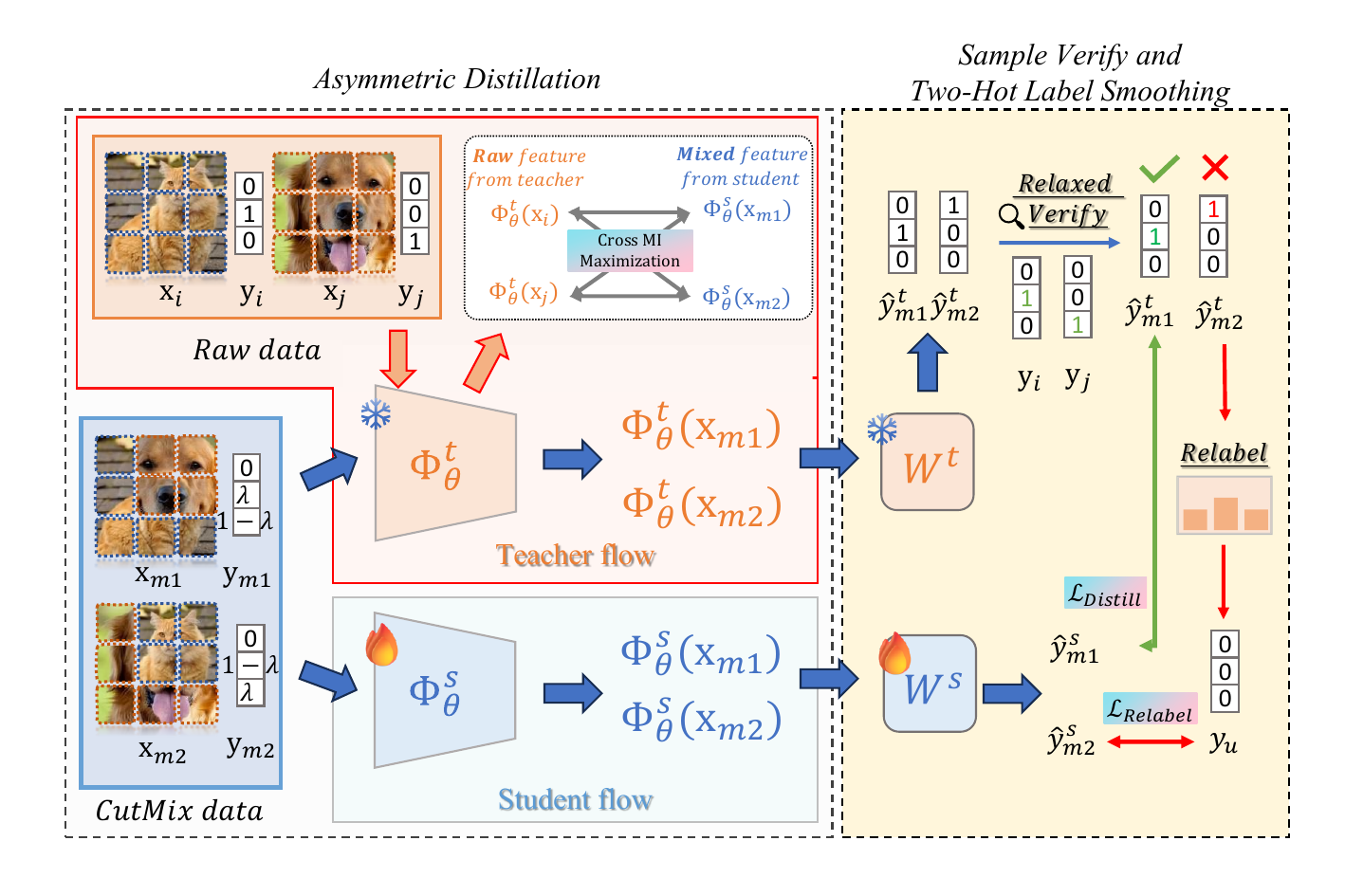}}
\caption{
The proposed asymmetric distillation framework.
Both the student and teacher models receive mixed data as input and perform distillation on $\Phi_\theta(x)$.
Besides, the teacher model additionally accepts raw data as input to enlarge its benefit on the mixed inputs.
To further decrease the student's activation of the non-discriminative features, we filter the teacher's wrong predictions of the mixed samples out and assign them a revised label to optimize.
}
\label{fig:pipeline_new}
\end{figure}

%% file: AnonymousSubmission/Sections/5_Experiments.tex
%实验
\section{Experiments}
\label{sec:experiments}
\subsection{Experimental Settings}
\input{AnonymousSubmission/Tables/main_results_total}
\input{AnonymousSubmission/Tables/SSB_result}
\noindent \textbf{Datasets.} We evaluate the performance of our model on three benchmarks: the OSR benchmark, semantic shift benchmark, and large-scale benchmark.
\begin{itemize}
\item \textbf{OSR Benchmark:} within this benchmark, the method is evaluated on five datasets, including SVHN~\cite{37648}, CIFAR-10~\cite{krizhevsky2009learning}, CIFAR+10, CIFAR+50, and Tiny-ImageNet~\cite{le2015tiny}. All settings align with those of AGC~\cite{vaze2022openset}.
\item  \textbf{Semantic Shift Benchmark:}  this evaluation protocol includes three datasets: Caltech-UCSD-Birds (CUB)\cite{wah2011caltech}, Stanford Cars\cite{Krause_2013_ICCV_Workshops}, and FGVC-Aircraft~\cite{maji2013fine}. The presence of specific attributes distinguishes different classes, and the difficulty of recognition is calculated based on the differences in the number of attributes.  Consequently, the open-set classes of the FGVC datasets are divided into `Easy', `Medium', and `Hard' levels to denote their similarities with close-set classes.   
\item  \textbf{Large-Scale Benchmark:} Within this protocol, 200 classes from Tiny-ImageNet are used for training. Subsequently, non-overlapping `Easy' and `Hard' splits from Imagenet-21k are selected for evaluation, following the approach outlined by Ren~\etal~\shortcite{ren20232}.
\end{itemize}

\noindent  \textbf{Evaluation Metrics.}
In the OSR benchmark, AUROC serves as a threshold-independent performance evaluation metric~\cite{davis2006relationship}. 
It quantifies the probability that a positive example possesses a higher detector score or value compared to a negative example. 
OSCR is a metric that gauges the trade-off between accuracy and open-set detection rate by adjusting the threshold on the confidence of the predicted class.

\noindent  \textbf{Implementation Details.} We utilize DIST~\cite{huang2022class} as the foundational distillation method. 
The default teacher-student pair consists of ResNet-50~\cite{he2016deep} and ResNet-18. 
The training duration spans 200 epochs, employing a batch size of 32. 
The initial learning rate is 0.1, subsequently reduced by a factor of 5 at the 60th, 120th, and 160th epoch. 
The optimization employs the SGD optimizer with a momentum of 0.9, and the weight decay is set to 5e-4.
    
\subsection{Comparison on OSR Benchmark}
To assess the recognition capability of our proposed asymmetric distillation framework in open-set scenarios, we compare it not only with traditional state-of-the-art open-set recognition methods (denoted as ARPL~\cite{chen2021adversarial}, RCSSR~\cite{huang2022class} and AGC~\cite{vaze2022openset}, etc.) but also with method CMPKD~\cite{wang2022what}, which incorporates both MSA and distillation.
As shown in Table~\ref{tab:auroc_total}, our method exhibits a significant improvement in open-set performance compared to traditional open-set recognition methods, achieving a gain of 0.4\% on CIFAR-10 and 2.6\% on TinyImageNet. 
In comparison to method CMPKD, our model substantially enhances the model's open-set performance while maintaining closed-set accuracy, achieving improvements of 4.9\% on SVHN and 2.8\% on TinyImageNet. 
This demonstrates that through the use of the asymmetric distillation framework and the constraints of mutual information object engineering, we have indeed succeeded in enhancing the focus of the student model on class-specific features, and finally got a win-win solution for open-set recognition.

\subsection{Comparison on Semantic Shift Benchmark}
To further explore the discriminative ability of our model for feature extraction, we conduct experiments on the semantic shift benchmark following AGC~\cite{vaze2022openset}. 
The results, as shown in Table~\ref{tab:SSB}, reveal that our method consistently outperforms the AUROC metric of the state-of-the-art method AGC by a margin of 1\%$\sim$2\% in both `Easy' and `Hard' splits while maintaining closed-set accuracy on three fine-grained datasets (CUB, SCars, and FGVC), a slightly declined by less than 1\% in the `Easy' split of the Aircraft, possibly due to the invariable backgrounds (either the sky or the runway) among the dataset. 
Notably, our model's performance excels in the hybrid scenario with an OSCR metric exceeding 2.7\% for the SOTA method. 
This suggests that, by employing MSA samples through an asymmetric distillation framework, our model can discard class-agnostic representations and focus more on class-specific representations, thereby enhancing recognition performance across various scenarios, even in challenging fine-grained classification scenarios.

\subsection{Comparison on Large-Scale Benchmark}
To further explore the effectiveness of our method in real-world scenarios, we conducted experiments on a large-scale dataset. 
Specifically, we trained on TinyImageNet with only 200 classes and tested on ImageNet-21k with 2100 classes. 
In our evaluation, we compare our method not only with conventional OSR methods, ARPL~\cite{chen2021adversarial} and AGC~\cite{vaze2022openset}, but also with additional multimodal methods such as CLIP~\cite{radford2021learning}, CoOp~\cite{zhou2022learning}, and A$^2$Pt.
The results, as presented in Table~\ref{tab:imagenet}, showcase the performance of our method in both the open-set recognition metric AUROC and the hybrid recognition metric OSCR. Remarkably, our method outperforms conventional methods by approximately 5\%, as well as multimodal methods by 1\%. 
This demonstrates that even in complex real-world scenarios, the features learned through the asymmetric distillation framework remain highly discriminative. 
Importantly, these features are not significantly disturbed by the increase in unknown novel classes, showcasing the ability of our method to stabilize the open-set recognition performance of the model.
Additionally, we replace our ResNet-18 backbone with ViT and report the results.
Compared to A$^2$Pt and CLIP, the performance on the ImageNet-21k dataset shows the superiority of our method on the ViT backbone. 
And the comparison with our ResNet-18 proves that using a more powerful backbone model can reap better benefits.

\subsection{Comparison on the Light-weight Model}
We conduct additional experiments to assess the effectiveness of our model on lightweight networks in Table~\ref{tab:mv2}. 
In comparison to the state-of-the-art (SOTA) method AGC~\cite{vaze2022openset} implemented on MobileNet-V2~\cite{sandler2018mobilenetv2}, our proposed asymmetric distillation model demonstrates superior performance. 
Our model exhibits improvements in both closed-set accuracy (Acc.) with a margin of +3.24\% and open-set recognition AUROC with a margin of +2.64\% on the TinyImageNet dataset. 
These results indicate that our method is not constrained by the network parameters and remains effective even in lightweight networks.
\input{AnonymousSubmission/Tables/imagenet}
\input{AnonymousSubmission/Tables/mv2}
\input{AnonymousSubmission/Tables/ablation_distillation}
\input{AnonymousSubmission/Tables/hyper}
\subsection{Ablation Study}
We conduct an ablation analysis on the Tiny-ImageNet dataset, as shown in Table~\ref{tab:ablation_distillation}, to delve deeper into the effectiveness of different elements of our method.
The comparison between ResNet-18 with CutMix (a) highlights the significant positive impact of multiple samples-based augmentations on improving closed-set classification (+3.1\%). However, this improvement comes at the expense of a substantial reduction in the model's open-set recognition performance (-1.8\%).
The introduction of distillation methods partially mitigates the degradation of open-set performance (from 81.0\% to 82.5\%), but does not lead to improvement, less than 82.8\%.
However, when our Contrastive Mutual Information (CMI) objective and Smoothed Two-Hot Label method are introduced, the open-set recognition metrics AUROC of the model sequentially increase from 82.5\% to 84.8\% and 85.3\%. 
Although the closed-set classification metric (Acc.) slightly decreased by 1\% point and 1.5\% points, respectively, there are still +3\% improvements over the original ResNet. 
This demonstrates that our proposed CMI objective and Smoothed Two-Hot Label method significantly enhance the model's open-set recognition ability. 
The empirical evidence highlights the effectiveness of our additional supervision methods in facilitating the learning of class-specific features and decreasing the activation of the non-salient features of the known classes.

To validate the robustness of our method to the hyper-parameters, we test different combinations of hyper-parameters $\mu$ and $\eta$ including 0.5, 1.0, and 2.0. 
The result in Table~\ref{tab:hyper} fluctuating around 0.5\% under different combinations shows that our method is insensitive to hyper-parameter settings. 
And the optimal result appears when $\mu$ and $\eta$ are set as 1.0.

\subsection{Results on Other Tasks}
\input{AnonymousSubmission/Tables/ood_results}
Our proposed method ensures the backbone model extracts features with discrimination and hence further promotes downstream task performance like OSR task. 
In Table~\ref{tab:ood_exp}, we evaluate our method on uncertainty-related task Out-of-Distribution (OoD) to verify our method as a general feature-strengthening tool. 
We equipped the maximum logit score (MLS) baseline with our method on Cifar-10/Cifar-100 and Cifar-10/Tiny-ImageNet and the improvements show the effectiveness on OoD task.

\input{AnonymousSubmission/Tables/medmnist}
Furthermore, we focus on the fundamental and practically applicable recognition task. As a win-win solution for close- and open-set tasks, our proposed method can be regarded as an
effective feature extractor enhancement strategy. 
Taking as an example, we verify the representation ability enhancement of our method on medical image analysis (on MedMNIST v2 dataset~\cite{yang2023medmnist}) in Table~\ref{tab:medmnist}.

%% file: AnonymousSubmission/Tables/main_results_total.tex
% \usepackage{multirow}
\begin{table*}[]
\centering
\caption{Comparison of AUROC (\%) and close-set accuracy (Acc., \%) on OSR Benchmark. The best performance values are highlighted in bold and the second best performances are underlined.}
\label{tab:auroc_total}
\renewcommand\arraystretch{1.2}
% 0.95
\scalebox{0.9}{
\begin{tabular}{lcc|cc|cc|cc|cc}
\hline
\toprule[2pt]
\multirow{2}{*}{Methods} & \multicolumn{2}{c|}{SVHN} & \multicolumn{2}{c|}{CIFAR-10} & \multicolumn{2}{c|}{CIFAR+10} & \multicolumn{2}{c|}{CIFAR+50} & \multicolumn{2}{c}{TinyImageNet} \\ \cline{2-11} 
                         & Acc. & AUROC & Acc. & AUROC & Acc. & AUROC & Acc. & AUROC & Acc. & AUROC \\ \hline
CROSR (CVPR, 2019)       & - & 89.9 & - & 88.3 & - & 91.2 & - & 90.5 & - & 58.9 \\
C2AE (CVPR, 2019)        & - & 92.2 & - & 89.5 & - & 95.5 & - & 93.7 & - & 74.8 \\
RPL (ECCV, 2020)         & - & 93.4 & - & 82.7 & - & 84.2 & - & 83.2 & - & 68.8 \\
ARPL+CS (TPAMI, 2021)    & - & 96.7 & - & 91.0 & - & 97.1 & - & 95.1 & - & 78.2 \\
CSSR (TPAMI, 2022)       & - & \textbf{97.9} & - & 91.3 & - & 96.3 & - & 96.2 & - & 82.3 \\
AGC (ICLR, 2022)         & 97.6 & 97.1 & 96.4 & \underline{93.6} & 97.8 & \underline{97.9} & 97.8 & \underline{96.5} & 84.6 & \underline{82.7}\\
OpenMix+ (TCSVT, 2023)   & - & - & 95.3 & 86.9 & 96.8 & 93.1 & 96.8 & 92.5 & 58.4 & 75.1\\\hline
CMPKD (NIPS, 2022)       & \underline{97.6} & 92.1 & \underline{96.8} & 84.1 & \underline{97.8} & 95.0 & \underline{97.8} & 91.9 & \underline{86.9} & 82.5            \\ \hline
Ours                     & \textbf{97.7} & \underline{97.2}     & \textbf{96.9}& \textbf{94.0} & \textbf{98.0}& \textbf{98.1} & \textbf{98.0}& \textbf{96.8} & \textbf{87.3}  & \textbf{85.3} \\
\bottomrule[2pt]
\end{tabular}}

\end{table*}

%% file: AnonymousSubmission/Tables/SSB_result.tex
% \usepackage{multirow}
\begin{table*}[]
\renewcommand\arraystretch{1.5}
\centering
\caption{Comparison of AUROC (\%), OSCR (\%) and closed-set accuracy (Acc., \%) on semantic shift Benchmark. 
Results of ARPL and AGC are from Vaze~\etal~\protect\shortcite{vaze2022openset}.
The best performance values are highlighted in bold.
}
\label{tab:SSB}
% 0.9
\scalebox{0.85}{
\begin{tabular}{lccc|ccc|ccc}
\hline
\toprule[2pt]
\multirow{2}{*}{Method} & \multicolumn{3}{c|}{CUB}          & \multicolumn{3}{c|}{SCars}        & \multicolumn{3}{c}{FGVC-Aircraft} \\ \cline{2-10} 
                        & \multirow{2}{*}{Acc.} & AUROC       & OSCR        & \multirow{2}{*}{Acc.} & AUROC       & OSCR        & \multirow{2}{*}{Acc.}  & AUROC       & OSCR        \\ 
                        &      & Easy / Hard & Easy / Hard &      & Easy / Hard & Easy / Hard &       & Easy / Hard & Easy / Hard  \\ \hline
ARPL (TPAMI, 2021)      & 85.9 & 83.5 / 75.5 & 76.0 / 69.6 & 96.9 & 94.8 / 83.6 & 92.8 / 82.3 & 91.5  & 87.0 / 77.7 & 83.3 / 74.9 \\
AGC (ICLR, 2022)        & 86.2 & 88.3 / 79.3 & 79.8 / 73.1 & \textbf{97.1} & 94.0 / 82.2 & 92.2 / 81.1 & \textbf{91.7}  & \textbf{90.7} / 82.3 & \textbf{86.8} / 79.8 \\
Ours                    & \textbf{87.6} & \textbf{89.6} / \textbf{82.0} & \textbf{81.4} / \textbf{75.8} & 96.9 & \textbf{95.5} / \textbf{84.9} & \textbf{93.3} / \textbf{83.5} & 91.1 & 90.1 / \textbf{83.5} & 86.2 / \textbf{80.9} \\  
% Ours(Teacher-free)      &  &  /  &  /  &  &  /  &  /  &  &  /  &  /  \\ 
\bottomrule[2pt]
\end{tabular}}

\end{table*}

%% file: AnonymousSubmission/Tables/imagenet.tex
\begin{table}[t]
\centering
\caption{Comparison of AUROC (\%) and OSCR (\%) on large-scale benchmark.}
\label{tab:imagenet}
\renewcommand\arraystretch{1.5}
\scalebox{0.7}{
\begin{tabular}{lc|cc|cc}
\toprule[2pt]
                       & \multirow{2}{*}{BackBone} & \multicolumn{2}{c|}{Easy} & \multicolumn{2}{c}{Hard} \\ \cline{3-6} 
                       & & AUROC       & OSCR       & AUROC       & OSCR       \\ \hline
ARPL                   & VGG32    & 51.4\%      & 33.8\%     & 52.4\%      & 34.4\%     \\
AGC                    & VGG32    & 72.8\%      & 44.1\%     & 72.1\%      & 43.8\%     \\ \hline
CLIP                   & ViT-B/32 & 72.9\%      & 44.0\%     & 72.3\%      & 43.6\%     \\
CoOp                   & ViT-B/32 & 74.6\%      & 54.3\%     & 73.3\%      & 53.3\%     \\
A$^2$Pt                & ViT-B/32 & 76.6\%      & 58.7\%     & 74.7\%      & 57.4\%     \\ \hline
Ours                   & ResNet18 & \underline{77.1\%}      & \underline{60.8\% }    & \underline{75.7\%  }    & \underline{60.1}\%     \\ 
Ours                   & ViT-B/32 & \textbf{79.3\%}      & \textbf{64.8\% }    & \textbf{75.7\%  }    & \textbf{62.8}\%     \\ \bottomrule[2pt]
\end{tabular}}
\end{table}

%% file: AnonymousSubmission/Tables/mv2.tex
\begin{table}[]
\renewcommand\arraystretch{1.5}
\centering
\caption{Comparison of AUROC (\%) and close-set accuracy (Acc., \%) on light-weight model MobileNet-V2.}
\label{tab:mv2}
\scalebox{0.9}{
\begin{tabular}{l|c|cc}
\toprule[2pt]
Method & Backbone & Acc. & AUROC \\ \hline
AGC & VGG-32 & 84.64\% & 82.68\%\\
AGC & MobileNet-V2 & 82.46\% & 80.68\%  \\
Ours & MobileNet-V2 & 85.70\% & 83.32\% \\ \bottomrule[2pt]
\end{tabular}}

\end{table}

%% file: AnonymousSubmission/Tables/ablation_distillation.tex
% \usepackage{multirow}
\begin{table}[]
\centering
\caption{Ablations of our proposed terms. All experiments are conducted on Tiny-ImageNet. The final close-set accuracy (Acc., \%) and AUROC (\%) are reported.}
\label{tab:ablation_distillation}
\renewcommand\arraystretch{1.2}
\scalebox{0.65}{
\begin{tabular}{lcccccc}
\toprule[2pt]
% \hline
Model                      & CutMix         & Distillation & CMI            & Smoothed two-hot label & Acc. & AUROC  \\ \hline
ResNet-18                  &                &              &                &                        & 84.3\%     & 82.8\% \\
(a)                        & \checkmark     &              &                &                        & 87.4\%     & 81.0\% \\
(b)                        & \checkmark     & \checkmark   &                &                        & 88.8\%     & 82.5\% \\
(c)                        & \checkmark     & \checkmark   & \checkmark     &                        & 87.8\%     & 84.8\% \\
% (d)                       & \checkmark     & \checkmark   &                & \checkmark             &      & \% \\ 
(d)                        & \checkmark     & \checkmark   & \checkmark     & \checkmark             & 87.3\%     & 85.3\% \\ % \hline
\toprule[2pt]
\end{tabular}}

\end{table}

%% file: AnonymousSubmission/Tables/hyper.tex
\begin{table}[]
\centering
\caption{Results under different hyper-parameter settings.}
\renewcommand\arraystretch{1.2}
\scalebox{0.8}{
\begin{tabular}{l|cc|cc|cc}
\toprule[2pt]
\multirow{2}{*}{\diagbox[dir=NW]{$\eta$}{$\mu$}} & \multicolumn{2}{c|}{0.5} & \multicolumn{2}{c|}{1.0} & \multicolumn{2}{c}{2.0} \\ \cline{2-7} 
                  & Acc.       & AUROC      & Acc.       & AUROC      & Acc.       & AUROC      \\ \hline
0.5               & 87.7\%           & 85.1\%           & 87.6\%           & 85.0\%           & 87.4\%           & 85.0\%           \\
1.0               & 87.3\%           & 84.9\%           & 87.3\%           & 85.3\%           & 87.5\%           & 85.1\%           \\
2.0               & 86.8\%           & 84.9\%           & 87.3\%           & 84.7\%           & 87.5\%           & 84.9\%           \\ \bottomrule[2pt]
\end{tabular}}
\label{tab:hyper}
\end{table}

%% file: AnonymousSubmission/Tables/ood_results.tex
% \usepackage{multirow}
\begin{table}[]
\centering
\caption{The evaluations on Cifar-10/Cifar-100 and Cifar-10/Tiny-ImageNet OoD detection.}
\renewcommand\arraystretch{1.2}
\scalebox{0.8}{
\begin{tabular}{l|c|c|c}
% \hline
\toprule[2pt]
Method & In-distribution           & Out-of-distribution        & AUROC \\ \hline
MLS    & \multirow{2}{*}{Cifar-10} & \multirow{2}{*}{Cifar-100} & 87.5\% \\
Ours   &                           &                            & 89.6\% \\ \hline
MLS    & \multirow{2}{*}{Cifar-10} & \multirow{2}{*}{Tiny-ImageNet} & 88.7\% \\
Ours   &                           &                            & 91.2\% \\ % \hline
\bottomrule[2pt]
\end{tabular}}
\label{tab:ood_exp}
\end{table}

%% file: AnonymousSubmission/Tables/medmnist.tex
% \usepackage{multirow}
\begin{table}[]
\centering
\caption{Results on MedMNIST v2 dataset.}
\renewcommand\arraystretch{1.2}
\scalebox{0.75}{
\begin{tabular}{lcc|cc|cc}
% \hline
\toprule[2pt]
\multirow{2}{*}{Method} & \multicolumn{2}{c|}{ChestMNIST} & \multicolumn{2}{c|}{OCTMNIST} & \multicolumn{2}{c}{PneumoniaMNIST} \\ \cline{2-7} 
                        & AUROC & Acc. & AUROC & Acc. & AUROC & Acc. \\ \hline
MedMNIST                & 76.8\% & 94.7\% & 94.3\% & 74.3\% & 94.4\% & 85.4\% \\
Ours                    & 77.5\% & 94.8\% & 96.2\% & 77.0\% & 94.9\% & 90.1\% \\ \bottomrule[2pt] % \hline
\end{tabular}}
\label{tab:medmnist}
\end{table}

%% file: AnonymousSubmission/Sections/6_Conclusion.tex
%结论
\section{Conclusion}
\label{sec:conclusion}
    In this paper, we start by revealing the two sides of the data-mix augmentation by investigating how MSA interplays with open-set recognition.
    Our experiments and visualizations suggest that MSA diminishes the criteria of OSR and leads to a confusion among the similar classes.
    Based on the observations of how knowledge distillation works on OSR, we propose a win-win solution which leverages MSA to boost both the close-set and the open-set performance.
    The outstanding performance of our method conducted on multiple datasets demonstrates the effectiveness of our approach.
    It also demonstrates the potential of the known classes can help to detect novels.

%% file: AnonymousSubmission/Sections/2_RelatedWork.tex
%相关工作
\section{Related Work}
\label{sec:relatedwork}
\subsection{Data Augmentation}
Data augmentation is a simple yet efficient way to improve the model's robustness and accuracy~\cite{simard2003best,devries2017dataset}.
Plain operations such as translation, rotation, flipping, and cropping can drastically enhance the model's classification competence.
The combinations of those operations with well-designed strategies were experimented~\cite{lemley2017smart,hendrycks2020augmix}, showing that models can benefit a lot from stronger augmentation operations.
Further, techniques that can shift the data distribution to encourage the model to learn more useful semantic contents and better classify the images were investigated to explore the potential of the models.
Additionally, the impact of the augmentations on models' uncertainty and robustness was investigated by Chun~\etal~\shortcite{chun2020empirical}.
Despite the enhancement of the expressive and generalization capabilities of the network through DA, they can lead to changes in the distribution of output features, thereby counterproductively affecting performance in OSR.
In this study, we further investigate the reasons behind the counterproductive impact of data augmentation on OSR, aiming to preserve the win-win performance of DA in both closed-set and OSR.

\subsection{Open-Set Recognition} 
Open-Set recognition methods handle challenges posed by samples from unseen classes during training.
Scholars employ various criteria to threshold the model's output information to estimate the likelihood of encountering unknown classes.   Prior works often utilize confidence-based and distance-based methods for this purpose.
The max logits~\cite{vaze2022openset} or the probabilities~\cite{scheirer2012toward,bendale2016towards,neal2018open} produced by the model can be seen as how many class-related features are detected in the input sample.
So threshold can be a straightforward way to pick the samples which the model is unfamiliar with.
However, the bottleneck remains that we need to train a classifier that can extract compact features and filter the class-agnostic features to reject the unknowns.
Some of the other methods~\cite{chen2021adversarial,cevikalp2023anomaly} evaluate the unfamiliarity by measuring the distance between the test samples and their learned latent representation.
The samples that are encoded far away from the cluster centers of the seen classes are judged to be unknown class samples.
Due to the unseen classes are totally unpredictable, the first priority is to make the most of what we can acquire during training~\cite{vaze2022openset}.
In this paper, we utilize the augmented data to further mine the potential of DA samples of seen classes and get a stronger open-set classifier.

%% file: IJCAI_2024/Sections/Analysis.tex
\section{Visualizations on MNIST Dataset}
    In Section 3.2 and Figure~2 of the main text, we discuss the class-wise similarities between known and unknowns.
    For a more intuitive view, we perform feature space visualizations using various models on the MNIST dataset as shown in Figure~\ref{fig:vis}, where the four unknown classes are partitioned in purple and six known classes are represented by other colors.
    Compared to the clusters of the vanilla model in Figure~\ref{fig:vis} (a), the dispersed features of the CutMix-trained model in Figure~\ref{fig:vis} (b) result in severe confusion among the classes.
    Our visual analysis substantiates the existence of the dispersion phenomena not only within closed-set classes but also among open-set classes, leading to confusion with the known classes.
    Notably, for each individual class in the feature space, the clusters associated with the vanilla model exhibit tighter cohesion compared to those of the CutMix-trained model.
    As shown in Figure~\ref{fig:vis} (c), our method restricts the unknown classes into the center of the feature space and enlarges both the inter-class margins and the intra-classes tightness to get a distinct boundary of all the classes.
    \input{AnonymousSubmission/Figures/fig_visualize}

%% file: AnonymousSubmission/Figures/fig_visualize.tex
\begin{figure}[!ht]
\centering

\includegraphics[width=1.0\linewidth]{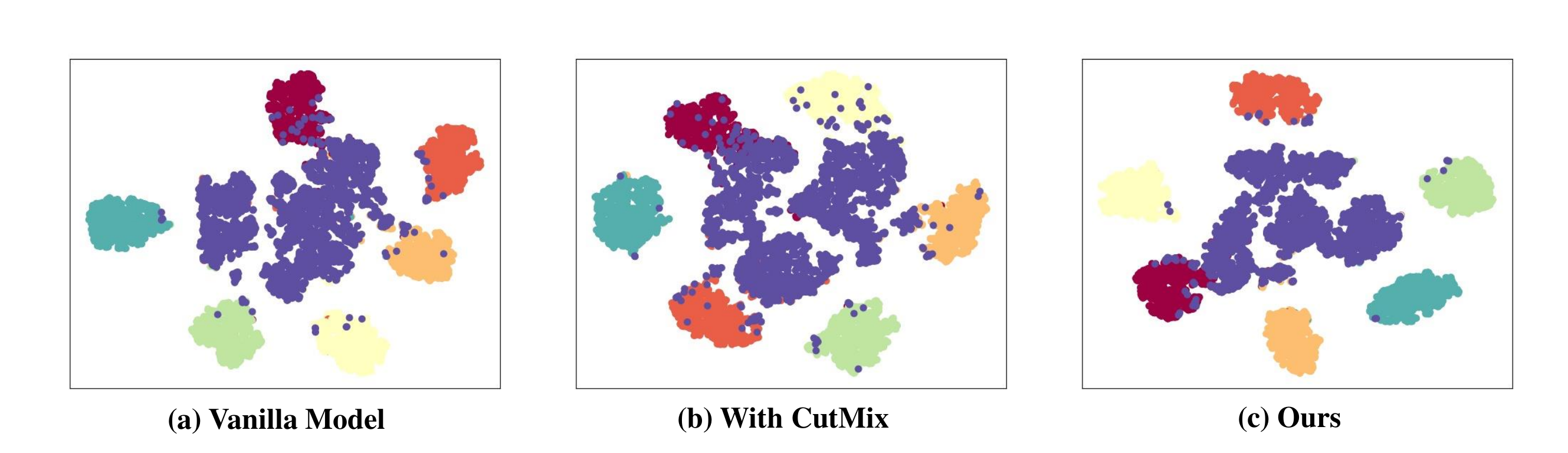}
\caption{Visualizations of the model's feature space on MNIST dataset. The purple points denote the unknown samples and the rest colors represent the known classes. (a) Vanilla CNN model. (b) CNN model with CutMix augmentation. (c) CNN model with our training framework.}
\label{fig:vis}
\end{figure}

%% file: IJCAI_2024/Sections/Extension.tex
\section{Extension for Teacher-free Learning}
    The asymmetric distillation framework outperforms previous works in our experiments.
    To achieve effective learning, we implement a lightweight teacher-free edition of our method by thresholding the predictions of the mixed samples and removing the teacher model.
    Since the over-confident predictions indicate sufficient information for classification and the uncertain predictions suggest ambiguous semantics, we pick them out to perform the mutual information maximization and the two-hot label smoothing separately.
    The loss function for this part consists of three loss terms: standard cross entropy loss for raw samples, cross mutual information loss, and uni-label loss for mixed samples.
   
    \noindent\textbf{Mix-batch Generation.}
    We train the raw samples and the mixed samples within the same batch to provide the objective of mutual information maximization.
    To achieve a diversity of mixing, we set the amount of the mixed samples two times the initial ones in our setting and applied different mixing coefficients within a mini-batch.

    \noindent\textbf{Upper Threshold for MI Maximization.}
    We omit the generated labels and apply the mutual information maximization to those mixed samples.
    Concretely, we first set an upper threshold $\tau_{upper}$ to filter out the mixed samples in which the model is quite confident with their predictions.
    The picked samples contain sufficient discriminative features so that we encourage those features to dominate the model's activation of the raw sample by mutual information maximization with the model's representation of the raw sample:
    \begin{equation}
    \begin{split}
        \mathcal{L}_{MI} = \mathds{1}&(\hat{\text{y}}_{m,c_1}> \tau_{upper})\,(- \mathcal{I}\,(\Phi_\theta(\text{x}_m), \Phi_\theta(\text{x}_i)))\\
        & + \mathds{1}(\hat{\text{y}}_{m,c_2}> \tau_{upper})\,(-\mathcal{I}\,(\Phi_\theta(\text{x}_m), \Phi_\theta(\text{x}_j))),
        % &+ (1 - \lambda)\,\mathcal{I}\,(\Phi_\theta(\text{x}_m), \Phi_\theta(\text{x}_j)))).
    \end{split}
    \end{equation}
    where $c_1$ and $c_2$ are the ground-truth classes of $\text{x}_i$ and $\text{x}_j$ and $\mathds{1}(\cdot)$ is an indicator function whose value is 1 when the following expression in the brackets is true and 0 vise versa.
    The upper threshold $\tau_{upper}$ is set to 0.95 in our experiments.

    \noindent \textbf{Lower Threshold for Uncertainty Learning.}
    % Figure~\ref{fig:top2_error} shows that model tends to make over-confident predictions for the mixed samples and some mixed samples can confuse the model to misclassify them.
    The upper threshold can enable the model to make full use of the mixed samples which includes the discriminative features for close-set classification, conversely, the confusing mixed samples can be utilized to elevate the model's ability for open-set detection.
    Specifically, we set a lower threshold $\tau_{lower}$ to pick out the model's predictions with high uncertainty.
    Then, we assign the picked-out samples with a uniformly distributed object $\bar{\text{y}}$ to maximize the entropy of their predictions.
    \begin{equation}
            \mathcal{L}_{Relabel} = \mathds{1}(\textup{max}(\hat{\text{y}}_m)<\tau_{lower})\,\mathcal{L}_{CE}(\hat{\text{y}}_m, \bar{\text{y}}).
    \end{equation}

    The overall object for the teacher-free learning is:
    \begin{equation}
            \mathcal{L} = \mathcal{L}_{CE} + \mu\,\mathcal{L}_{MI} + \eta\,\mathcal{L}_{Relabel},
    \end{equation}
    where $\mu$ and $\eta$ are hyper-parameters we set to 1.0.

    \noindent\textbf{Results.}
    Since the teacher-free framework introduces the two extra thresholds for the different objectives, we conduct experiments in Table~\ref{tab:diff_thre} to validate the robustness of our method under various thresholds.
    % In Table~\ref{tab:diff_thre}, we experiment with our teacher-free framework on Tiny-ImageNet dataset under various thresholds.
    In comparison to the vanilla CNN, our method improves both the accuracy and the AUROC of the model under eight different combinations of the thresholds.
    % our method improves both the accuracy and the AUROC under different thresholds, which demonstrates its robustness.
    The optimal values of the upper and the lower thresholds are 0.8 and 0.5, which reaches the best performance on both accuracy and AUROC.
    
    In Table~\ref{tab:ablation_ssl}, we draw ablations of the teacher-free edition of our method.
    Three experiments are conducted: (a) utilizing only mutual information objective, (b) only thresholding the ambiguous ones to learn uncertainty, and (c) our complete teacher-free edition.
    From this table, we observe that both (a) and (b) can retrieve the degradation of the model's open-set performance (+2.6\%) with a lighter decrease in the model's classification accuracy (-0.5\% and -1.3\%).
    This also demonstrates the effectiveness of the two terms of our method.
    The mutual information objective enables the model to learn more classification information to perform better on close-set performance to help the model reach a higher accuracy.
    By combining both the two terms, we can achieve a better open-set performance (+3.0\%) and alleviate the degradation of the model's accuracy in (b).
    However, the final performance is not as good as the asymmetric distillation framework, which indicates that the asymmetric distillation framework is also essential for the win-win solution.
    \input{IJCAI_2024/Tables/diff_thre}
    \input{IJCAI_2024/Tables/ssl_ablation}

%% file: IJCAI_2024/Tables/diff_thre.tex
% \usepackage{multirow}
\begin{table}[]
\centering
\renewcommand\arraystretch{1.2}
\scalebox{0.5}{
\begin{tabular}{cc|l|cc|cc|cc|cc}
\hline
\toprule[2pt]
\multicolumn{2}{c|}{Vanilla CNN} & \multirow{2}{*}{
\diagbox[dir=NW]{Lower}{Upper}}
& \multicolumn{2}{c|}{0.7} & \multicolumn{2}{c|}{0.8} & \multicolumn{2}{c|}{0.9} & \multicolumn{2}{c}{0.95} \\ \cline{1-2} \cline{4-11} 
                              Acc. & AUROC &  & Acc.       & AUROC       & Acc.       & AUROC       & Acc.       & AUROC       & Acc.       & AUROC       \\ \hline
\multirow{2}{*}{84.3} & \multirow{2}{*}{82.8} & 0.5 & 85.96      & 83.55       & \textbf{86.44} & \textbf{84.03} & 85.98   & 83.98   & 85.82      & 83.96       \\
                      &                       & 0.6 & 85.92      & 83.73       & 86.28      & 83.76       & 85.90      & 83.67       & 85.76      & 83.67       \\ \hline
\toprule[2pt]
\end{tabular}}
\caption{Performances of the teacher-free framework on Tiny-ImageNet dataset under various thresholds. We report the AUROC (\%) and accuracy (Acc., \%).}
\label{tab:diff_thre}
\end{table}

%% file: IJCAI_2024/Tables/ssl_ablation.tex
% \usepackage{multirow}
\begin{table}[]
\centering
\renewcommand\arraystretch{1.2}
\scalebox{0.7}{
\begin{tabular}{lcccc}
% \hline
\toprule[2pt]
Model       & MI         & Uni-label    & Acc. & AUROC \\ \hline
ResNet-18   &  &  & 87.4 & 81.0 \\
(a)         & \checkmark &  & 86.9 & 83.7 \\
(b)         &  & \checkmark & 86.1 & 83.7 \\
(c)         & \checkmark & \checkmark & 86.4 & 84.0 \\ % \hline
\toprule[2pt]
\end{tabular}}
\caption{Ablations for the teacher-free framework. All experiments are conducted on Tiny-ImageNet and the final AUROC (\%) and accuracy (Acc., \%) are reported.}
\label{tab:ablation_ssl}
\end{table}

%% file: IJCAI_2024/Sections/Experiments.tex
\section{More Experimental Results}
% \section{Supplementary Results}

\subsection{More Results on OSR Benchmark}
    Despite CutMix, we combine our asymmetric distillation framework with different augmentations, including MixUp, Decoupled-CutMix~\cite{liu2022decoupled} for MSA and CutOut for SSA on OSR benchmark.
    The results in Table~\ref{tab:sup_more_results_osr_benchmark} show that our method works under different augmenting techniques.
    For CutOut, we only use the mutual information objective during the optimization.
    
    % We mainly use the CutMix augmentation in our paper.
    % Without loss of generality, more augmentations are introduced into our asymmetric distillation framework on OSR benchmark to further investigate the effectiveness of our method.
    % For multiple-sample-based augmentation, we add the experiments with MixUp and Decoupled-CutMix.
    % Table~\ref{tab:sup_more_results_osr_benchmark} shows the results.
    % Except for the multiple-sample-based augmentation, we try exerting our method on CutOut to get a further gain on model's open-set performance.

    MSAs show their powerful abilities on the model's open-set performance.
    By combining different MSA techniques with our method, both the classification accuracy and the open-set performance of the model are enhanced.
    This also confirms the universality of our method of eliminating the degradation of open-set performance in MSA.
    Additionally, combing our method with CutOut can also reach a comparable performance comparing to MSA.
    % Jointed with CutOut, our method brings the model a further improvement on AUROC and can get the comparable performance comparing to MSA.
    
    % The multiple-sample-based augmentations bring the model better close-set performance.
    % Moreover, the open-set performance can also be improved when combing the augmentations with our approach.
    % Among the results, the CutMix augmentation shows its superior performance.
    % This also confirms the universality of our method on eliminating the degradation of open-set performance in MSA.
    
    % We select CutOut to validate the effectiveness of our method on single-sample-based augmentation.
    % The mutual information maximization objective of CutOut is the teacher's prediction of its raw sample.
    % We do not exert relaxed verification on this experiment.
    % Since the single-sample-based augmentation can slightly improve the model's open-set performance, our asymmetric distillation framework can further boost the model's performance.
    % The performance of CutOut even surpasses our method on Cifar and SVHN datasets.
    % The results show that our approach can also perform well with various augmentations.
 
    \input{IJCAI_2024/Tables/diff_aug}

\subsection{Visualization of Uncertainty Distribution}
    \input{AnonymousSubmission/Figures/fig_distribute}
    Following Cen~\etal\shortcite{jun2023devil}, we visualize the model's uncertainty by calculating $1-max(\hat{\text{y}})$ on the test set of Tiny-ImageNet dataset in Figure~\ref{fig:distribute}.
    The blue bar indicates the uncertainty of known samples, and the orange bar denotes the unknown classes.
    Open-set recognition aims to predict the known classes with low uncertainty and the unknown classes with higher uncertainty.
    In comparison to Figure~\ref{fig:distribute} (b), the CutMix-trained model in Figure~\ref{fig:distribute} (a) predicts all classes with higher uncertainty so that the model can not discriminate the unknown classes.
    The distilled model in Figure~\ref{fig:distribute} (b) retrieves the model's low uncertainty of known classes.
    However, it tends to predict the unknown classes with low uncertainty in the reason of the increased feature norm for all classes.
    In Figure~\ref{fig:distribute} (c), our method achieves a trade-off between the high uncertainty of the CutMix-trained model for unknown classes and the low uncertainty of the distilled model for known classes by selectively increasing the feature norm of the discriminative features.
\subsection{More Ablations}
    \input{IJCAI_2024/Tables/ablation}

    \textbf{Ablations of Asymmetric Distillation Framework.}
    We investigate the effectiveness of various micro designs within each component in Table~\ref{tab:sup_abaltion}.
    For example, we consider the class in a larger proportion as the one-hot label of the mixed sample.
    Then, the relaxed verification of the teacher's predictions can be replaced by an exact one in which we check the teacher's top-1 prediction to filter out the wrong predictions.
    The wrong predictions are then relabeled with the two-hot label smoothing.
    Furthermore, in Table~\ref{tab:sup_abaltion} (b), `Single MI maximization' denotes that the objective of the mutual information maximization is the raw sample that takes a larger part in the mixture, and we omit the optimization of another raw sample.
    
    Filtering the teacher's predictions in a relaxed manner enables the model to optimize with more mixed samples with the teacher's guidance.
    The comparison of (a) and (c) highlights the advantage of the relaxed verification in filtering the predictions on the model's open-set performance (from 84.7\% to 84.9\%).
    With the single mutual information maximization, the model gets a higher close-set performance (+0.3\%) at the cost of the degradation of open-set performance (-0.4\%).
    Our complete method (c) retains more information within the mixed samples and balances close-set and open-set performance well.

%% file: IJCAI_2024/Tables/diff_aug.tex
% \usepackage{multirow}
\begin{table}[!t]
\renewcommand\arraystretch{1.3}
\centering
\scalebox{0.5}{
\begin{tabular}{lcc|cc|cc|cc|cc}
\hline
\toprule[2pt]
\multirow{2}{*}{Method} & \multicolumn{2}{c|}{SVHN}          & \multicolumn{2}{c|}{Cifar-10}      & \multicolumn{2}{c|}{Cifar+10}      & \multicolumn{2}{c|}{Cifar+50}      & \multicolumn{2}{c}{Tiny-ImageNet} \\ \cline{2-11} 
                       & Acc. & AUROC & Acc. & \multicolumn{1}{c|}{AUROC} & Acc. & \multicolumn{1}{c|}{AUROC} & Acc. & \multicolumn{1}{c|}{AUROC} & Acc.            & AUROC           \\ \hline
CNN+CutOut             & 97.3 & 97.0 & 96.3 & 93.6 & 97.7 & 97.7 & 97.7 & 96.3 & 86.4 & 84.2 \\
Ours+CutOut            & 97.8 & 97.3 & 96.9 & 94.0 & 98.0 & 98.2 & 98.0 & 97.0 & 86.7 & 84.5 \\ \hline
CNN+MixUp              & 98.3 & 92.8 & 97.6 & 92.7 & 98.7 & 91.2 & 98.7 & 85.3 & 86.8 & 82.6 \\
Ours+MixUp             & 97.6 & 97.2 & 96.8 & 93.4 & 97.9 & 97.5 & 97.9 & 96.1 & 87.1 & 84.7 \\ \hline
CNN+DCutMix            & 98.5 & 90.1 & 98.1 & 82.5 & 98.6 & 88.4 & 98.6 & 83.2 & 87.6 & 80.0 \\ 
Ours+DCutMix           & 97.7 & 97.3 & 96.8 & 93.6 & 98.0 & 97.5 & 98.0 & 96.0 & 88.1 & 84.6 \\ \hline
CNN+CutMix             & 98.1 & 93.4 & 97.5 & 83.5 & 98.6 & 91.2 & 98.6 & 85.5 & 87.4 & 81.0 \\ 
Ours+CutMix            & 97.7 & 97.2 & 96.9 & 94.0 & 98.0 & 98.1 & 98.0 & 96.8 & 87.8 & 85.3 \\ \bottomrule[2pt]
\end{tabular}
}
\caption{Results of our method with different augmentations on OSR benchmark. We report the AUROC (\%) and accuracy (Acc., \%).}
\label{tab:sup_more_results_osr_benchmark}
\end{table}

%% file: AnonymousSubmission/Figures/fig_distribute.tex
\begin{figure}[!ht]
\centering
\scalebox{1}{
\includegraphics[width=1\linewidth]{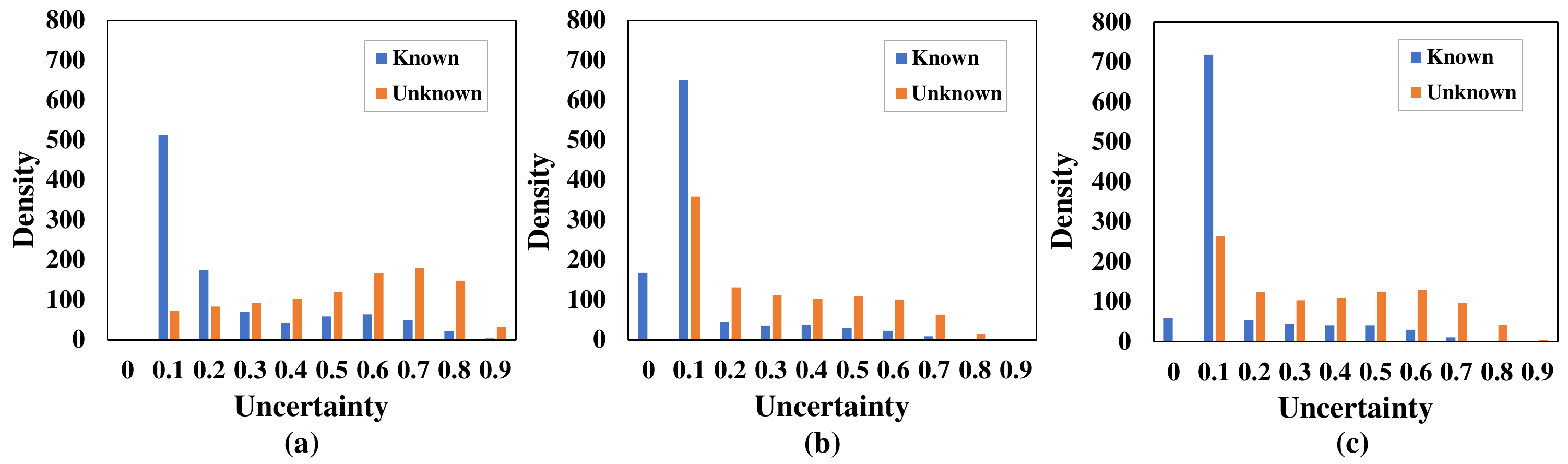}}
\caption{
The uncertainty distribution visualizations for different method on Tiny-ImageNet dataset.
(a) Vanilla CutMix training framework.
(b) Symmetric distillation framework.
(c) Our asymmetric distillation framework.
} 
\label{fig:distribute}
\end{figure}

%% file: IJCAI_2024/Tables/ablation.tex
% \usepackage{multirow}
\begin{table}
\centering
\renewcommand\arraystretch{1.3}
\scalebox{0.8}{
\begin{tabular}{lclclcc}
\hline
\toprule[2pt]
\multirow{2}{*}{Method}    & \multicolumn{2}{c}{Verification} & \multicolumn{2}{c}{MI Maximization} & \multirow{2}{*}{Acc.} & \multirow{2}{*}{AUROC} \\ \cline{2-5}
                           & \multicolumn{1}{l}{Exact} & Relaxed & \multicolumn{1}{l}{Single} & Cross &  & \\ \hline
ResNet-18                  &  &  &  &  & 87.4 & 81.0  \\
(a)                        & \multicolumn{1}{c}{\checkmark} &  &  & \multicolumn{1}{c}{\checkmark} & 87.4 & 84.7  \\
(b)                        & & \multicolumn{1}{c}{\checkmark} & \multicolumn{1}{c}{\checkmark} &  & 87.6 & 84.9   \\
(c)                        & & \multicolumn{1}{c}{\checkmark} &  & \multicolumn{1}{c}{\checkmark} & 87.3 & 85.3   \\ \bottomrule[2pt]
                           % & & \multicolumn{1}{c}{\checkmark} &                                  & \multicolumn{1}{c}{\checkmark} & \multicolumn{1}{c}{\checkmark} & 84.2                   \\ \bottomrule[2pt]
                           % &                                 &                                  &                                &                                &                    \\
                           % &                                 &                                  &                                &                                &                    \\ \bottomrule[2pt]
\end{tabular}
}
\caption{Ablations of various micro designs within relaxed verification and mutual information maximization. All experiments are conducted on Tiny-ImageNet and the final AUROC (\%) is reported.}
\label{tab:sup_abaltion}
\end{table}